\documentclass{article}
\pdfoutput=1 

\usepackage[nonatbib, preprint]{neurips_2023}

\usepackage[utf8]{inputenc} 
\usepackage[T1]{fontenc}    
\usepackage{hyperref}       
\usepackage{url}            
\usepackage{booktabs}       
\usepackage{amsfonts}       
\usepackage{nicefrac}       
\usepackage{microtype}      

\usepackage{microtype}
\usepackage{graphicx}
\usepackage{booktabs} 

\usepackage[dvipsnames]{xcolor}

\usepackage{amsmath}
\usepackage{amssymb}
\usepackage{mathtools}
\usepackage{amsthm}

\usepackage{amsmath,amsfonts,bm}

\def\eqref#1{equation~\ref{#1}}

\def\1{\bm{1}}

\DeclareMathAlphabet{\mathsfit}{\encodingdefault}{\sfdefault}{m}{sl}
\SetMathAlphabet{\mathsfit}{bold}{\encodingdefault}{\sfdefault}{bx}{n}

\newcommand{\R}{\mathbb{R}}

\theoremstyle{plain}
\newtheorem{theorem}{Theorem}[section]

\newtheorem{lemma}[theorem]{Lemma}

\theoremstyle{definition}

\theoremstyle{remark}

\usepackage[textsize=tiny]{todonotes}

\usepackage{hyperref}
\usepackage{url}
\usepackage{xspace}
\usepackage{booktabs}
\usepackage[normalem]{ulem}
\usepackage{enumitem}
\usepackage{amsmath}
\usepackage{graphicx}
\usepackage{amssymb}
\usepackage{wrapfig}
\usepackage{multirow}
\usepackage{subfig}
\usepackage{caption}

\def\@fnsymbol#1{\ensuremath{\ifcase#1\or *\or \dagger\or \ddagger\or
   \mathsection\or \mathparagraph\or \|\or **\or \dagger\dagger
   \or \ddagger\ddagger \else\@ctrerr\fi}}
\newcommand{\ssymbol}[1]{^{\@fnsymbol{#1}}}

\newcommand{\method}{WavSpA\xspace}
\newcommand{\paradigm}{WavSpA\xspace}
\newcommand{\methodname}{Transformer-AdaWavSpA\xspace}

\newcommand{\diff}{\mathop{}\!\mathrm{d}}
\newcommand{\Lm}{\mathop{}\!\mathbb{L}}

\title{WavSpA: Wavelet Space Attention for Boosting Transformers' Long Sequence Learning Ability}

\author{%
  Yufan Zhuang \\
  UC San Diego\\
  \And
  Zihan Wang \\
  UC San Diego \\
  \And
  Fangbo Tao \\
  Mindverse \\
  \And
  Jingbo Shang \\
  UC San Diego \\
}

\begin{document}
\maketitle

\begin{abstract}

    Transformer and its variants are fundamental neural architectures in deep learning. 
    Recent works show that learning attention in the Fourier space can improve the long sequence learning capability of Transformers. 
    We argue that wavelet transform shall be a better choice because it captures both position and frequency information with linear time complexity.
    Therefore, in this paper, we systematically study the synergy between wavelet transform and Transformers.
    We propose \underline{Wav}elet \underline{Sp}ace \underline{A}ttention (\paradigm) that facilitates attention learning in a learnable wavelet coefficient space which replaces the attention in Transformers by (1) applying forward wavelet transform to project the input sequences to multi-resolution bases, (2) conducting attention learning in the wavelet coefficient space, and (3) reconstructing the representation in input space via backward wavelet transform. 
    Extensive experiments on the Long Range Arena demonstrate that learning attention in the wavelet space using either fixed or adaptive wavelets can consistently improve Transformer's performance and also significantly outperform learning in Fourier space. We further show our method can enhance Transformer's reasoning extrapolation capability over distance on the LEGO chain-of-reasoning task.
\end{abstract}

\section{Introduction}


Transformer~\cite{vaswani2017attention} has become one of the most influential neural architectures in deep learning. Large language models such as ChatGPT~\cite{ChatGPT} have reshaped people's imagination of what an AI model can do in making conversation with humans, solving nontrivial math problems, writing code, and even co-authoring a paper~\cite{kung2022performance}. In image processing, vision transformers have become the backbone for a wide array of applications~\cite{dosovitskiy2020image, radford2021learning}. Similarly, on source code understanding, Codex~\cite{codex} can finish people's code given the helper text of the function or just the function name. All of those accomplishments are built upon the foundational Transformer.

Nevertheless, the effective handling of long sequences remains a challenge for Transformers due to the intricate relationships that can exist within such sequences. 
To address this limitation, recent research has focused on enhancing the Transformers' long-range capabilities through attention learning in transformed sequence spaces. 
One approach involves low-cost token-mixing, which utilizes forward Fourier transformation to achieve notable accuracy improvements while maintaining quasi-linear time complexity~\cite{fnet}. 
However, without incorporating a backward transformation, the model might inadvertently mix information from both the input and transformed spaces. 
To overcome this limitation, researchers have leveraged the forward and backward Fourier transformations to learn large filters with linear weights~\cite{GFNET} and non-linearities~\cite{guibas2021adaptive} for vision tasks, exploiting the equivalence between multiplication in the Fourier space and direct convolution in the input space.


\begin{figure*}[t]
\centering
\vspace{-5mm}
\includegraphics[width=\linewidth, keepaspectratio]{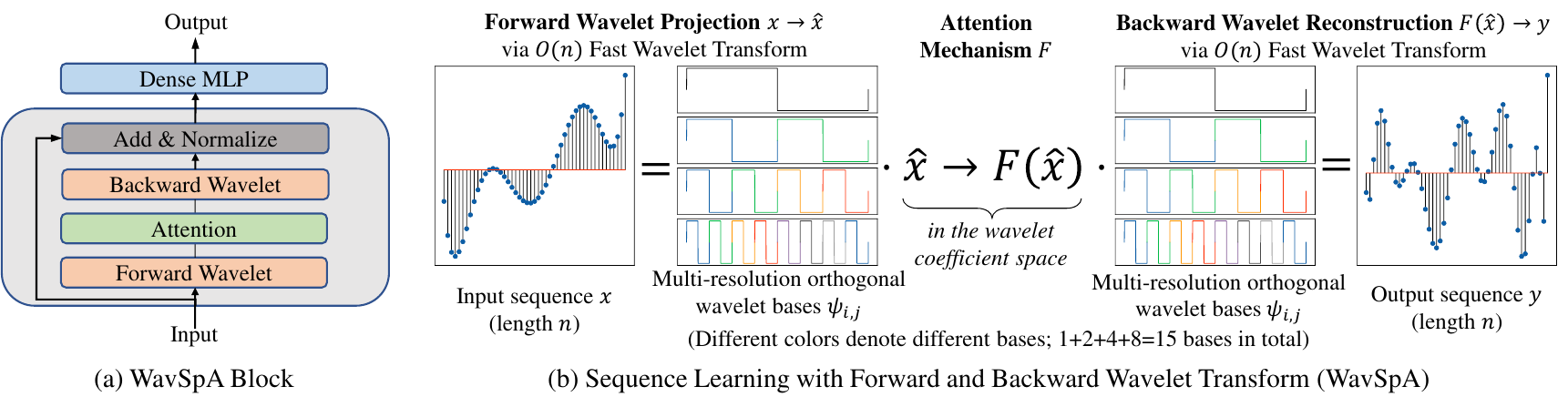}
\caption{An overview of our proposed \paradigm. (a) The only difference between a Transformer block and a \paradigm block is the attention computation. 
(b) The general flow of computation in \paradigm with learnable forward and backward wavelet transform.}
\label{fig:overview}
\vspace{-6mm}
\end{figure*}

In light of these developments, it is evident that attention learning in transformed sequence spaces holds significant promise for enhancing the effectiveness of Transformers' handling of long-range dependencies.
We propose \underline{Wav}elet \underline{Sp}ace \underline{A}ttention (\paradigm) that facilitates attention learning in a learnable \emph{wavelet coefficient space}, as shown in Figure~\ref{fig:overview}(a).
Specifically, we first apply \emph{forward} wavelet transform to project the input sequence to multi-resolution bases, then conduct attention (e.g., full attention~\cite{vaswani2017attention}, random feature kernel~\cite{rahimi2007randomkernel})
in the wavelet coefficient space, and finally, reconstruct the representation in input space via \emph{backward} wavelet transform.
We implement the transform using Fast Wavelet Transform (FWT)~\cite{mallat1989theory} so both transform steps are linear in time, leading to a small overhead.

Performing attention on a sequence in a wavelet-transformed space can offer several advantages. 
Firstly, it can enhance the representation of the input sequence by capturing relevant features and patterns. By applying the transformation, the sequence is mapped to a new space where certain characteristics might be easier to capture. Attention mechanisms can then be applied in this transformed space to effectively weigh these transformed features, leading to improved representation learning.
Secondly, it can enable the attention mechanism to capture different types of relationships between the elements of the sequence, such as associative relationships. By operating in the transformed space, attention can effectively capture the underlying structure of the data and reason over it, leading to improved performance on long sequences.
Finally, it is orthogonal to existing work that attempts to replace attention, hence can be combined with any Transformer design.

Besides applying fixed wavelets, we further propose three ways to construct learnable wavelets: direct wavelet parameterization, orthogonal wavelet parameterization, and wavelet lifting. We give detailed explanations of the three schemes and discuss their individual advantages and drawbacks.


We conduct extensive experiments on the Long Range Arena (LRA) benchmark to validate and justify our proposed \paradigm. 
By combining fixed wavelet space with various representative attention methods, we observed significant performance improvements without introducing additional time complexities.
Furthermore, we analyze the performance of \paradigm's three parameterization schemes when coupled with the attention methods, demonstrating even stronger performance boosts.
Additionally, our investigation demonstrated that equipping the Transformer with our proposed \paradigm resulted in enhanced reasoning extrapolation capacity, as evidenced by improved performance on the LEGO dataset~\cite{LEGO}.
These findings highlight the superior long-range understanding capabilities achieved by learning in the wavelet coefficient space compared to the input space or Fourier space.


In summary, our major contributions are as follows.
\begin{itemize}[leftmargin=*,nosep]
    \item We propose \paradigm to facilitate learning in the wavelet space following a forward-backward paradigm which can be paired with various attention methods and boost their long-range understanding capabilities. 
    \item We further propose three adaptive wavelet parameterization schemes (Ada\paradigm, Ortho\paradigm, Lift\paradigm) to maximize the flexibility of wavelet transformation. 
    \item Extensive experiments on the Long-Range Arena benchmark have demonstrated the effectiveness and also justified the design of \paradigm. 
    \item We show \paradigm  enhances the reasoning extrapolation capacity to longer sequence lengths. 
\end{itemize}
\noindent\textbf{Reproducibility.} We will release our code on GitHub.


\section{Learning Attention in a Transformed Space}
Inspired by recent work, we begin our study with sequence space transformation with Fourier transforms. 
FNet~\cite{fnet} replaced the attention with solely forward Fourier transform, it performs well empirically but mixing Fourier coefficients with the input of the original data space is not an intuitive approach. 
Typical space transforms consist of a forward step and a backward step~\cite{GFNET, guibas2021adaptive}.
Hence, we are interested in comparing sequence learning in a forward-only or in a forward-backward mode.

We conduct pilot studies on the Text task of Long Range Arena~\cite{lra}, combining various attention mechanisms with Forward Only Fourier transform or Forward Backward Fourier transform. 
The results are summarized in Table~\ref{tab:ablation2}, and experiment details can be found in Section~\ref{sec:experiment}. Notably, we observed that learning with the Forward Backward mode consistently outperformed the Forward Only mode. While the Fourier transform occasionally outperformed the original space, its improvement was not consistently observed across all attention mechanisms.


\begin{table}
    \centering
    \small
    \begin{minipage}{\textwidth}   
    \centering
    \caption{
    Transformed Spaces vs. Original Space (N/A) on the Long Range Arena Text task. 
    We color the number green if it surpasses the baseline (i.e., N/A), red vice versa.}    \label{tab:ablation2}
    \begin{tabular}{l ccccc} 
      \toprule
      Transformation & Transformer & Linformer & Linear Att. & Longformer & Performer \\
      \midrule
      Original Space (N/A) & 64.27 & 53.94 & 65.90 & 62.85 & 65.40 \\
      \midrule
      Fourier - Forward Only~\cite{fnet} & \textcolor{purple}{54.65} & \textcolor{purple}{51.27} & \textcolor{purple}{65.25} & \textcolor{purple}{53.51} & \textcolor{purple}{53.39} \\
      Fourier~\cite{GFNET, guibas2021adaptive} & \textcolor{purple}{56.42} & \textbf{\textcolor{ForestGreen}{57.06}} & \textcolor{ForestGreen}{71.66} & \textcolor{purple}{55.36} & \textcolor{ForestGreen}{65.52} \\ 
      Fixed Daubechies-2 Wavelet & \textbf{\textcolor{ForestGreen}{74.82}} & \textcolor{ForestGreen}{55.22} & \textbf{\textcolor{ForestGreen}{71.93}} & \textbf{\textcolor{ForestGreen}{74.99}} & \textbf{\textcolor{ForestGreen}{75.60}} \\
      \bottomrule
    \end{tabular}
   \end{minipage}
\end{table}

This phenomenon is understandable since Fourier transform maps signals into the frequency domain, resulting in the loss of time information.
In the deep learning context, losing time information is analogous to losing positional information. And positional information is vital in many tasks, as it pins down associative relationships amid elements of the sequence. 
Hence, preserving and leveraging time information becomes vital for effectively capturing the dependencies within the sequence.

\begin{figure}
\centering
\includegraphics[width=\columnwidth]{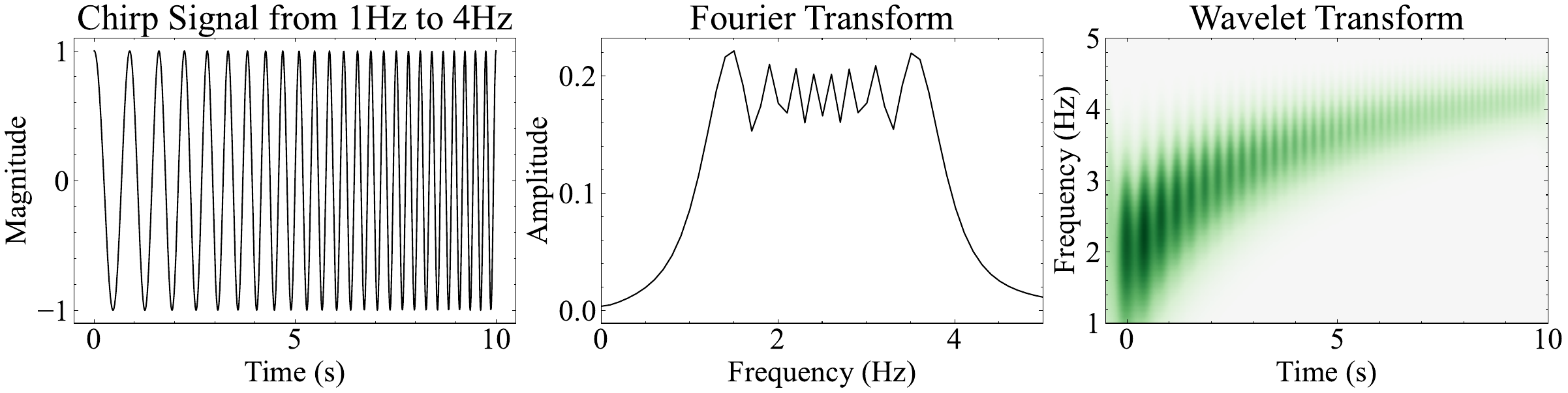}
\vspace{-3mm}
\caption{We show a chirp signal from 1Hz to 4Hz, its continuous Fourier transform, and its continuous wavelet transform. 
From the Fourier transform graph one can only infer the existence of signal in the range of 1-4Hz without time information, while in the wavelet transform graph, both time and frequency information are present and one can tell this is a chirp signal.}
\label{fig:FT_WT}
\vspace{-7mm}
\end{figure}

Based on such observation, we propose \paradigm that facilitates attention learning in a wavelet coefficient space, detailed methodology explained in Section~\ref{sec:methodology}. Wavelet transform is a sequence projection method where both frequency and time information are captured. As an illustration, we show an example of wavelet transform to demonstrate its ability in time-frequency localization compared to the Fourier transform (see Figure~\ref{fig:FT_WT}). Furthermore, the wavelet transform is multi-level where the decomposition levels correspond to low-to-high frequencies. In the deep learning context, low-frequency signal represents global features and high-frequency signal represents local features, which has been shown useful in prior attention methods~\cite{longformer, bigbird}. 

This multi-level decomposition capability corresponds to the multi-level nature of long inputs such as human text. 
As associative relationships in text occur at various levels, starting from individual words within a sentence. For instance, in the sentence``\emph{The cat chased the mouse}'' the words ``\emph{cat}'' and ``\emph{mouse}'' are associated in terms of their roles in the action.

Associative relationships also extend beyond sentence boundaries. Texts are organized in hierarchical structures, such as paragraphs, sections, and documents, where higher-level associations emerge. Within a paragraph, sentences are associated, contributing to a coherent idea. In longer texts like news articles, sections and chapters form hierarchical connections, uniting them under common topics.

This hierarchical structure is not unique to text but also exists in other sequential inputs, including source code, formulas, and more. Recognizing and understanding this multi-level hierarchy is crucial as it enables models to capture rich relationships within the sequence, facilitating more advanced extrapolation reasoning capabilities.




To validate our intuition, we perform experiments on the LRA benchmark (Fixed Daubechies-2 Wavelet row of Table~\ref{tab:ablation2}), the results indicate wavelet transform can deliver consistent performance boosts across a wide range of attention mechanisms. Furthermore, we present a comprehensive comparison of attention learning in Fourier space and Fixed wavelet spaces in Appendix Table~\ref{tab:main_appendix}. 


\section{\method: Learning Attention in Parametrized Wavelet Space}
\label{sec:methodology}
In this section, we introduce the details of \method. 
As shown in Figure~\ref{fig:overview}(a), the only difference between a Transformer block and a \method block is the attention computation. 
The general flow of \paradigm is shown in Figure~\ref{fig:overview}(b), which constitutes the forward wavelet transform, the attention in the middle, and the backward wavelet transform. 

We list our notations here --- we denote scalars as $x$, vectors as $\mathbf{x}$, matrices as $X$; 
we denote function $f$'s transformation in the coefficient space as $\hat{f}$.


\subsection{\paradigm Paradigm}
We propose the \paradigm paradigm to conduct attention learning in the wavelet coefficient space between forward and backward transformation. 
The forward transformation decomposes the input sequence into coefficients of a set of wavelet basis.
We then conduct attention in the coefficient space. 
In the backward transformation, we reconstruct the target representation in the original function space.
For fixed wavelet families, we require the forward-backward transformation pair to be invertible and exact, meaning that one can perfectly reconstruct the same input from the derived coefficients. However, this constraint is not always attached to adaptive wavelets. 

The general framework is shown below. In practice, we deal with vectors with dimensions of the attention head dimension. Here, we limit ourselves to 1d functions for a clear illustration. 
Given input and output function $x(t), y(t) : \R \xrightarrow{} \R$ on time domain $t$, wavelet basis $\psi(\omega, t)$ on both frequency and time domain $\omega,t$ (e.g, the basis for a Daubechies-2 wavelet), and attention module $\mathrm{Attention}$,  
\begin{align}
    \text{(forward)}\quad &\hat{x}(\omega) = \sum_{i} x(t_i) \psi^*(\omega, t_i)\\
    \text{(attention)}\quad &\hat{h}(\omega) = \vphantom{\frac11} \mathrm{Attention} \circ \hat{x}(\omega)  \\
    \text{(backward)}\quad &y(t) = \vphantom{\frac11} \sum_j \hat{h}(\omega_j) \psi(\omega_j, t)
\end{align}
where $\psi^*(\omega, t)$ denotes the complex conjugate of $\psi$. 


Learning carried out in this space will correspond to gathering and processing information in a coarse to fine-grained fashion. Furthermore, wavelet transform enjoys $O(n)$ time complexity~\cite{mallat1989theory}, an already desirable property compared to Fourier transform's $O(n\log n)$ complexity.

\subsection{Direct Wavelet Parameterization - Ada\paradigm}
One key benefit of wavelet transformation is its flexibility in choosing the wavelets for its application, for example, Daubechies wavelets~\cite{daubechies1992ten} are optimized to have the most compact support; symlets~\cite{daubechies1988orthonormal} are designed to have better symmetric properties. 
Therefore it is natural to consider parameterization of the wavelet coefficients and make wavelet transformation part of the learning process.

The direct parameterization scheme is the most intuitive approach.
We make the wavelet coefficients learnable parameters, and update them during training.
The key problem here is maintaining the structure between the scaling coefficients and the wavelet coefficients, i.e. the quadrature mirror filter (QMF) relationship~\cite{daubechies1988orthonormal}.
We consider parameterizing the scaling coefficients ($\phi^{(n)} \in \mathbb{R}^{n}$, $n$ denotes wavelet length) and expanding the system according to the QMF relationship to obtain the full set of wavelet coefficients ($\psi^{(n)} \in \mathbb{R}^{n}$), shown in equation~\ref{eqn:qmf}.
\begin{align}
    \psi^{(n)}_{j} = (-1)^{j} \phi^{(n)}_{-j},\quad j \in \mathbb{Z} \label{eqn:qmf}
\end{align}
Further strengthening the learning power of adaptive parameterizations, we use different sets (i.e., $d$ sets) of learnable wavelets for individual hidden dimensions of the input $X\in \mathbb{R}^{n,d}$. At the same time, we do not wish the output to have volatile changes when permuting the hidden dimensions. In other words, we want permutation invariance for the hidden dimensions. 
For that reason we only use 1d wavelet transform over the input's hidden dimension $d$ for parameterized transformations, including this scheme and the following two parameterization schemes, 

The direct parameterization scheme satisfies the QMF relationship automatically, but we have no guarantee that the trained wavelet will form an orthogonal wavelet basis. We enjoy more freedom at the cost of using a potentially imperfect projection and reconstruction pair.

\subsection{Orthogonal Wavelet Parameterization - Ortho\paradigm}
We provide another way to systematically construct parameterized orthogonal wavelets to keep the perfect reconstruction property intact.
There exist extensive studies on this topic~\cite{vaidyanathan1988lattice, lina1993parametrizations, rieder1998parameterization, lattice1}, but many are based on constrained optimization, which is not ideal for our purpose. 
We present an unconstrained construction that originates from lattice filters, we refer readers to~\cite{lattice1} for details of this design.
In general, the orthogonal wavelets are constructed iteratively, each time we extend the wavelet by multiplying the current wavelet by an upshifted rotation matrix.
The resulting wavelet basis will always be orthogonal, the formula is shown below:
\begin{align}
    \psi^{(n)} = \mathrm{R}(\theta_1) \cdot \mathrm{U}  \cdot \ldots \cdot \mathrm{U} \cdot \mathrm{R}(\theta_n) 
\end{align}
where $\mathrm{R}$ is the rotation matrix and $\mathrm{U}$ is an upshift matrix.

As an example, we show how to construct a parameterized wavelet $\psi^{(4)}$ of length 4 from a parameterized wavelet of length 2 ($\psi^{(2)} = \left[ \sin{\theta_1}, \cos{\theta_1} \right]$):
\begin{align}
    \begin{bmatrix}
    \psi_{4}^{(4)} \\
    \psi_{3}^{(4)} \\
    \psi_{2}^{(4)} \\
    \psi_{1}^{(4)}
    \end{bmatrix} =  
    \begin{bmatrix}
    \cos_{\theta_2} & 0\\
    -\sin_{\theta_2} & 0\\
    0 & \sin{\theta_2}\\
    0 & \cos{\theta_2}
    \end{bmatrix}
    \begin{bmatrix}
    \sin{\theta_1} \\
    \cos{\theta_1}
    \end{bmatrix}
\end{align}
$\theta_1,\; \theta_2$ represent the two rotation angles that we can set as learnable parameters.

This parameterization scheme offers naturally orthogonal wavelets without the need to customize the loss functions or derive a new optimization process. But on the other hand, this scheme requires more computation than the direct parameterization scheme and the compute cost grows with respect to the wavelet length.

\subsection{Wavelet Lifting - Lift\paradigm}
The wavelet lifting scheme~\cite{sweldens1998lifting} is developed to become the second-generation wavelet transformation, due to its simplicity and extended flexibility. 
It is not characterized by transformation via functional convolution, rather it builds its forward and backward transformation from these three steps: 1. \textit{segmentation}: splitting the input into two parts, one widely used segmentation is separating the even and odd parts of the input; 2. \textit{update}: we mix the information from the subsampled segment into the wavelet segment  3. \textit{lifting}: normalizing the subsampled segment and blend the information again.

The simplest design is the so-called Lazy wavelet~\cite{sweldens1998lifting, lifting_tutorial}, the forward transformation is shown below for the first level where $(\lambda_{1,:}, \gamma_{1,:})$ represent the subsampled coefficients and wavelet coefficients:
\begin{align}
    \text{(segmentation)}\quad &\lambda_{1, k} = x_{2k},\; \forall k \in \mathbb{Z} \\
    \text{(update)}\quad &\gamma_{1,k} = x_{2k+1} - \frac{1}{2} (\lambda_{1,k} + \lambda_{1,k+1}),\; \forall k \in \mathbb{Z}\label{eqn:updater} \\
    \text{(lifting)}\quad &\lambda_{1, k} = \lambda_{1, k} + \frac{1}{4}(\gamma_{1,k-1}, \gamma_{1,k}),\; \forall k \in \mathbb{Z} \label{eqn:lift}
\end{align}
It is assumed that each point in the input is related to its neighbors, hence in equation~\ref{eqn:updater} we mix the information from the even segment to the odd segment. 
Then to make sure each decomposition level has the same mean and energy, we lift the subsampled coefficients with the wavelet coefficients, mixing the odd segment into the even segment in equation~\ref{eqn:lift}.
In the Lazy wavelet lifting scheme, the wavelets are inexplicitly parameterized by non-linearities they are later applied to.

A second-level decomposition ($\lambda_{2,:},\; \gamma_{2,:}$) will further decompose the $\lambda_{1,:}$ into finer-grained sequences. And the backward transformation is straightforward: simply reversing the positive and negative signs in the forward steps accordingly will recover the segments.

Wavelet lifting is a simple and straightforward alternative wavelet transformation scheme. 
The update and lifting step could be subject to arbitrary designs, which entitled this scheme with the most flexibility.
However, what comes with this flexibility is the huge search space for finding the optimal lifting, hence we only use the basic Lazy wavelet in our study and leave the rest for future research.



\section{Experiments}
\label{sec:experiment}
Our study begins by conducting experiments on the publicly available benchmark Long Range Arena~\cite{lra}. Our aim is to compare the effectiveness of learning attention in different input spaces: the regular input space, Fourier space with 2D Fourier transformation, and wavelet space with fixed 2D Daubechies-2 wavelet transformation. The results of these experiments demonstrate noteworthy improvements in performance, shown in Table~\ref{tab:ablation2}. 

Furthermore, we proceed to examine the performance of \paradigm's three parameterization schemes when combined with attention methods. The outcomes reveal even more substantial performance gains, as illustrated in Table~\ref{tab:adaptive_table}. Based on these findings, we propose a hypothesis that attributes these performance boosts to enhanced reasoning capabilities over distance. To validate our hypothesis, we test it on the LEGO~\cite{LEGO} dataset, which is a chain-of-reasoning task.

In addition to our primary investigations, we perform runtime analysis to measure the add-on cost imposed by~\paradigm, the result shows that the overhead is small (Appendix~\ref{sec:appendix_runtime}). We compare the performance of our adaptive \paradigm when coupled with attention (\methodname) with other efficient transformers on the LRA benchmark in (Appendix Table~\ref{tab:lra_table}). We provide a proof to show \paradigm maintains Transformer's universal approximation power (Appendix~\ref{sec:appendix_proof}). We also conduct ablation studies to verify the significance of backward reconstruction and the importance of wavelet initialization in the training process (Appendix~\ref{sec:ablation} ). In the end, we embark on an exploratory case study to investigate the characteristics of the learned wavelets (Appendix~\ref{sec:exploratory}).

\subsection{Experimental Design}
\textbf{Long Range Arena (LRA)} \quad 
LRA~\cite{lra} is designed to compare efficient transformers for their long-range reasoning ability. Since its release which already contains ten different efficient transformers, more and more efficient transformers have chosen it as the primary evaluation target. 
The datasets require understanding long sequences of mathematical operations, classifying text based on sentiment, matching similar documents, classifying images, and recognizing 2D spacial information. The sequence lengths of the dataset are within the range of 1K-4K. 

\textbf{LEGO} \quad
LEGO~\cite{LEGO} is a reasoning task that encapsulates the problem of following a chain of reasoning. The task itself requires reasoning over a sequence of variables and operators, and figuring out the sign of each variable. A sample input sequence will look like this: $a=+1;b=-a;e=+b;d=-f;c=+d;f=+e;$, and the model will be asked to predict the sign (positive/negative) of each variable. We follow the design of~\cite{LEGO}, train on the first 14/20 variables, and test on all 20/26 variables for our experiments. We train all models from random initialization.

\textbf{Wavelet Transformation Details} \quad 
For fixed \paradigm, we use Daubechies-2 wavelet that has length 4 as the default choice and apply 2d wavelet transform with one decomposition level over both the sequence length and hidden dimension. 
For adaptive \paradigm, we only transform over the sequence length axis because we intend to avoid large permutation variance over the hidden dimensions since we enabled learning distinctive adaptive wavelets over them. 
In our experiments, for direct wavelet parameterization we initialize from Daubechies-20 wavelet that has length of 40 or Daubechies-8 wavelet that has length of 16, for orthogonal wavelet parameterization we set the wavelet length as 16, for wavelet lifting we conduct three levels of decomposition. The detailed hyper-parameters are reported in Appendix~\ref{sec:appendix_lra}.
\begin{table}
  \begin{center}
    \caption{Evaluation results of our \paradigm paradigm with the three adaptive parameterization schemes. 
    We denote original space as N/A.
    Following previous works~\cite{luna, xiong2021nystromformer}, due to prolonged training on retrieval task, we also report 
    mean test accuracy without this task, denoted as ``(w/o r)''.
    }
    \small
    \resizebox{\textwidth}{!}{
    \begin{tabular}{c | cccc | cccc }
    \toprule
    \multicolumn{1}{c}{\multirow{2}{*}{\textbf{Model}}} & \multicolumn{4}{c}{\textbf{LRA Mean Test Acc}} & \multicolumn{4}{c}{\textbf{LRA Mean Test Acc (w/o r)}}  \\
    \cmidrule(lr){2-5} \cmidrule(lr){6-9} 
     & \textbf{N/A} & \textbf{Ada\paradigm} & \textbf{Orth\paradigm} & \textbf{Lift\paradigm} & \textbf{N/A} & \textbf{Ada\paradigm} & \textbf{Orth\paradigm} & \textbf{Lift\paradigm}  \\
    \midrule
    \textbf{Transformer} & 54.39 & \textbf{\textcolor{ForestGreen}{70.59}} & \textcolor{ForestGreen}{65.90} & \textcolor{ForestGreen}{59.85} &  53.62 & \textbf{\textcolor{ForestGreen}{68.43}} & \textcolor{ForestGreen}{64.50} & \textcolor{ForestGreen}{60.70}  \\
    \textbf{Linformer} & 49.36 & \textcolor{ForestGreen}{50.72} & \textcolor{ForestGreen}{52.01} & \textbf{\textcolor{ForestGreen}{52.12}} & 48.64 & \textcolor{purple}{48.12} & \textbf{\textcolor{ForestGreen}{49.95}} & \textcolor{purple}{47.47} \\
    \textbf{Linear Att.} & 50.67 & \textbf{\textcolor{ForestGreen}{64.32}} & \textcolor{ForestGreen}{55.86} & \textcolor{ForestGreen}{56.93} & 50.06 & \textbf{\textcolor{ForestGreen}{62.55}} & \textcolor{ForestGreen}{55.86} & \textcolor{ForestGreen}{57.65} \\
    \textbf{Longformer} & 53.46 & \textbf{\textcolor{ForestGreen}{63.66}} & \textcolor{ForestGreen}{54.96} & \textcolor{ForestGreen}{57.48} & 52.60 & \textbf{\textcolor{ForestGreen}{64.93}} & \textcolor{ForestGreen}{54.96} & \textcolor{ForestGreen}{58.54} \\
    \textbf{Performer} & 51.41 & \textbf{\textcolor{ForestGreen}{65.47}} & \textcolor{ForestGreen}{60.69} & \textcolor{ForestGreen}{56.95} & 50.81 & \textbf{\textcolor{ForestGreen}{64.05}} & \textcolor{ForestGreen}{61.44} & \textcolor{ForestGreen}{58.00} \\
    \bottomrule
    \end{tabular}
    }
    \label{tab:adaptive_table}
  \end{center}
  \vspace{-6mm}
\end{table}

\textbf{Experiment Environment}. \quad
Our early-stage experiments are conducted on RTX 3090 GPUs and later moved to TPU v2-8s and v3-8s. Our code is written in Jax~\cite{jax2018github} with the Flax framework~\cite{flax2020github}. The fixed wavelet transformation implementation is primarily based on Jax Wavelet Toolbox~\cite{wavelettoolbox} and PyWavelets~\cite{Lee2019}.

\subsection{Attention in Fixed~\paradigm}
Our \paradigm paradigm has a general philosophy of applying attention in the wavelet space and is not limited to a certain type of attention method. 
We comprehensively evaluate representative attention methods on different space transformations (no transformation, 2d Fourier transformation, and 2d fixed wavelet transformation with Daubechies-2 wavelet). 
In Table~\ref{tab:ablation2}, we show that performing full attention, or many other attention approximation operations in a wavelet transformed space as proposed in \paradigm paradigm almost always brings great accuracy improvements.  The complete result is shown in Appendix Table~\ref{tab:main_appendix}. Almost all attention methods have increased accuracy when trained in the wavelet space compared to an untransformed space or the Fourier space, except for the Image dataset, where some incur a slight drop in accuracy. 


\subsection{Attention in Adaptive~\paradigm}
We demonstrate that the parameterized wavelets can further boost attention methods' performance. In Table~\ref{tab:adaptive_table}, we show results for the three parameterization schemes mentioned in Section~\ref{sec:methodology} when each of these schemes is coupled with full attention and several other representative attention methods. The full result is included in Appendix Table~\ref{tab:adaptive_table_full}.

From the experiment results, we observe that direct parameterization almost always provides the highest accuracy elevation, followed by orthogonal parameterization and lifting. 
This is counter-intuitive: one would think imposing more structures should help the model to learn better wavelets, and in some cases it does, but our experiments show that learning wavelets with the most freedom is the best option most of the time.
Does this mean wavelets' nice mathematical properties are not essential at all, and any parameter initialization would work?

We conduct ablation studies where we initialize the directly parameterized wavelets from Gaussian distribution $\mathrm{N}(\mu=0, \sigma=0.02)$, and from damped sinusoidal waves ($x[t] = \frac{\cos(t)}{t+1}$).
The results are shown in Appendix Table~\ref{tab:ablation_init}. This showcases the importance of initializing from wavelets even when we impose no constraints on them.

\begin{figure}[ht]
   \begin{minipage}{0.45\textwidth}
     \centering
     \includegraphics[width=0.98\linewidth, keepaspectratio]{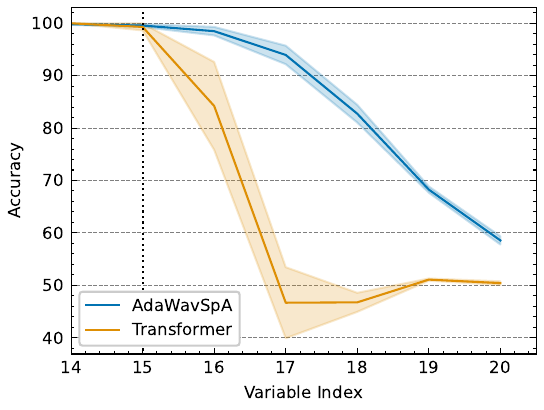}
   \end{minipage}\hfill
   \begin{minipage}{0.45\textwidth}
     \centering
     \includegraphics[width=0.98\linewidth, keepaspectratio]{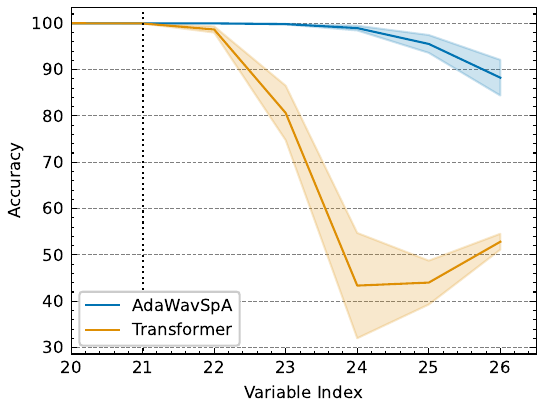}
   \end{minipage}
   \caption{Generalization of \methodname and vanilla Transformer. The \paradigm paradigm improves the reasoning extrapolation to longer sequence lengths. The two figures show the test set results for LEGO with 20 and 26 variables respectively. We report 90\% confidence intervals.} \label{fig:lego}
   \vspace{-3mm}
\end{figure}

\subsection{LEGO Reasoning Task}
We hypothesize that the observed performance gain comes from the enhanced reasoning power over distance. We test our hypothesis with the LEGO chain-of-reasoning task. Following the original configuration, the Transformer is a randomly initialized BERT-base model (12-layer, 768-hidden dimensions) and Ada\paradigm represents the same model when wrapped in our framework. 
We train on the first 14/20 variables and evaluate on the last 6 variables. The learning rate (5e-5), training schedule (200 epochs), and batch size (1024) are all following the original configuration. 
We perform three runs for each model, the result is shown in Figure~\ref{fig:lego}. It can be observed that the Transformer's extrapolation-over-distance capability is significantly enhanced when coupled with our framework.








\section{Related Work}
\label{sec:related}
\subsection{Attention Methods}
There has been plenty of prior work to enable transformers to handle long input more effectively and efficiently.  
Since the inefficiency comes from the quadratic dependency on sequence length because of the dense attention operation, a large portion of research simulates the attention operation with certain approximations, for example, replacing the dense attention matrix with a sparse version, or assume that it satisfies certain low-rank structures.
We briefly review some methods on this topic in this section. 
For a more detailed survey, we refer the readers to~\cite{efficient_survey}. 


\textbf{Sparse Attention}.\quad 
Perhaps the most intuitive solution to alleviate the quadratic cost, Sparse Attention only calculates a portion of the full $n^2$ attention matrix. 
Early stage methods include Local Attention~\cite{local_attention} and Multi-passage BERT~\cite{wang2019multi} use sliding windows or chunked blocks to speed up computation. 
Longformer~\cite{longformer} and BigBird~\cite{bigbird} further combine global attention, sliding window attention, dilated sliding window attention, and random attention together to form strong sparse attention mechanisms, and BigBird showed that their method is a universal approximator of sequence functions.  
On the other front, Orthogonal Transformer~\cite{huang2022orthogonal} utilizes an iterative approach to construct an orthogonal vector basis in Euclidean space, then perform windowed attention on grouped tokens after orthogonal projection. 

\textbf{Low-rank Approximation}. \quad
The self-attention matrix, at the center of transformer, has been found to display low-rank behaviors after pre-training.
Linformer~\cite{linformer} performed spectrum analysis on the pre-trained attention matrix, and the results indicate that the top 128 singular values composite 88\%-96\% of the entire 512 singular values across attention heads and layers.
Based on this observation, Linformer added low-rank projection matrices in attention to approximate the original attention matrix. 
On a similar notion, Drone~\cite{drone} extended the low-rank approximation scope to all matrices in transformer via data-driven optimal compression.

\textbf{Kernel Methods}. \quad
The kernel methods approximate the whole self-attention by replacing the softmax with a kernel function that can be decomposed to avoid the explicit calculation of the $O(n^2)$ matrix multiplication.
Linear Transformer~\cite{linear_transformer} proposed a non-negative $\mathrm{elu}$ feature mapping as the substitution for the softmax, they further pointed out the connection between their formulation and RNNs, and argued that transformers and RNNs can be unified under the same umbrella. 
Building on top of this, Random Feature Attention~\cite{random_feature_attention} and Performer~\cite{performer} utilized random feature approximation of the attention, one highlights the importance of normalization before random projection while the other one emphasizes the benefits of positive \& orthogonal random features.  

\textbf{Token Mixing}. \quad
Token Mixing methods are another version of efficient transformer building blocks. Different from the methods discussed above, they do not approximate attention, but rather conduct a new way of enabling communication between tokens. 
Hard-coded Gaussian attention~\cite{gaussian_averaging} showed the possibility that a random token mixing strategy can work well in transformer encoders, as opposed to delicate (pre-)trained attention heads. Token Mixing is a new view towards attention learning as these methods do not perform self-attention at all. FNet~\cite{fnet} pushed this idea further by providing an efficient method to mix the tokens with Fourier forward transformation. 

Among these methods, our \method utilizes wavelet transform, thus, is slightly similar to Token Mixing. However, our work should be seen as a new approach to boost transformers, which mixes the idea of a sequence space transform that communicates between tokens and attention methods that can benefit from the new space. 

\subsection{State Space Models}
Different from all the attention methods, state space models (SSM) such as S4~\cite{s4} construct long-term memories utilizing orthogonal polynomial projection. They update and maintain the hidden states according to a differential equation, and output the states using linear projection. They have shown outstanding performance on LRA and other long-range tasks. It is also flexible in choosing the family of orthogonal polynomials, but for each polynomial family (Laguerre, Legendre) and each measure (uniform, truncated), significant effort is required to derive the explicit SSM formula.
Similarly, MEGA~\cite{mega} utilized the exponential moving average mechanism to construct its hidden space for recording long-range dependencies and has shown promising results.
Our \paradigm is orthogonal towards the SSMs since our target is to boost attention methods' performance on long-range tasks as a sequence space transformation paradigm.

\subsection{Sequence Space Transformation in ML}
In the field of machine learning, sequence space transformations have gained widespread usage. One particularly common transformation is the Fast Fourier Transform (FFT)~\cite{fft1965}, which is frequently employed due to its ability to speed up convolution operations. In the context of our research objective, previous works such as AFNO~\cite{guibas2021adaptive} and GFNet~\cite{GFNET} have explored the learning of global filters for images by incorporating block-wise MLPs in the Fourier transformation process. This approach can be seen as akin to utilizing a convolutional layer with large filters. However, it is important to note that these methods were primarily designed for learning global filters. Through our comprehensive analysis (Table~\ref{tab:main_appendix}) and ablation study (Table~\ref{tab:ablation}), we have demonstrated that such architectures are inadequate for capturing long-range associative relationships.


Fast wavelet transform\cite{mallat1989theory} has been used for neural network compression~\cite{wave_compression}, speech recognition~\cite{tufekci2000feature}, and time series analysis~\cite{michau2022fully}. Recently in computer vision, WaveMix~\cite{Wavemix} proposed to mix the input images with forward wavelet transform. We note that our work differs from theirs by learning the attention in the coefficient space amid forward and backward wavelet transform.

\section{Conclusions and Future Work}

In this paper, we propose to learn attention in the wavelet coefficient space. Specifically, the inputs are first forward transformed into the wavelet space, then the attention is learned, and finally, we reconstruct the transformed sequence back in the input space. When coupled with attention methods, learning in wavelet space can boost their performance on long-range understanding tasks while enjoying no extra cost in time complexity. 
We further propose three ways to learn adaptive wavelets for \paradigm: direct parameterization, orthogonal parameterization, and wavelet lifting. We discuss their advantages and drawbacks and evaluate them empirically. The experiments show adaptive wavelets can provide an even stronger lift to attention methods.
In the end, we conduct study on a chain-of-reasoning task, to show the improved long-range learning capability may come from the enhanced reasoning extrapolation power. 


Through this work, we have focused on performing attention in the transformed wavelet space, either via fixed wavelet transformation or adaptive wavelet transformation. Is there an optimal way to construct learnable wavelets? And if so what should be the leading criterion for such optimality? These are both interesting questions we leave for the future.




\newpage
\section*{Acknowledgements}
Yufan Zhuang is supported by UC San Diego Jacobs School of Engineering Fellowship. We thank Sahil Suneja, Weitang Liu, Bill Hogan, Dheeraj Mekala, Yuwei Zhang, and Letian Peng for the valuable discussions.

\bibliography{bibliography}
\bibliographystyle{plain}

\appendix
\section{Appendix}

\subsection{Wavelet Transform} \label{sec:appendix_wavelet}
Fourier transform decomposes the entire function into global sinusoidal waves. 
It tells people what \emph{frequencies} are there and in what \emph{magnitude}, but no information is given about \emph{when} that frequency started or ended. 
See Figure~\ref{fig:FT_WT} for an illustration on a chirp signal.
This limits the capability to understand the local structures of the input and to conduct learning on top of it, which is crucial to many machine learning tasks. 


Wavelet transform is designed to solve this issue. 
We give a basic introduction here, we refer interested readers to~\cite{daubechies1992ten} for a more thorough explanation.
Wavelet transform employs a function $\psi(x), x \in \R$, called mother wavelet, to generate a family of translated and dilated wavelets (see Figure~\ref{fig:overview}(b)): 
\begin{align}
    \psi_{i,j}(x) = 2^{\frac{i}{2}} \psi(2^i x - j),\quad i,j \in \mathbb{Z}
\end{align}
where scale $i$ controls the resolution 
of the wavelet and $j$ controls the position of the wavelet. 
With a larger $i$ the wavelet will be squeezed shorter in space, hence the normalization factor $2^{\frac{i}{2}}$ to ensure the same $L^2$ norm for all wavelets.
The wavelet family $\psi_{i,j}(x)$ is orthogonal on this dyadic grid.

To be a valid mother wavelet $\psi(x)$, the only requirement is admissibility:
\begin{align}
    \int_{\R} \psi(x) \diff x = 0
\end{align}
In other words, the sum of 
function value should be 0. 

Given any square integrable function $f \in \mathbb{L}^2(\R)$ (i.e., $\int |f(x)|^2 \diff x < \infty $) and wavelet functions $\psi_{i,j}$, the wavelet transform pair is defined as:
\begin{align}
    &\hat{f}(i,j) = \int_{\R} f(x) \psi^*_{i,j}(x) \diff x = \sum_{t} f(x_t) \psi^*_{i,j}(x_t) \label{eq:dwtf}\\
    &f(x) = \sum_{i=-\infty}^{+\infty} \sum_{j=-\infty}^{+\infty} \hat{f}(i,j) \psi_{i,j}(x)  \label{eq:dwtb}
\end{align}
where $\psi^*_{i,j}(x)$ denotes the complex conjugate of $\psi_{i,j}(x)$.

Intuitively, in wavelet transform, we are scanning $f(x)$ with a microscope that has two knobs. 
One knob is the location $j$, the other one is the frequency (i.e., $2^i$).
We will be able to oversee the local structure of the input and calibrate it accordingly with parameterized functions in \paradigm paradigm. 

To generalize beyond $\Lm^2(\R)$ and avoid using an infinite number of wavelets, we must introduce another function $\phi$, called scaling function with a similar admissibility and orthogonality constraint:
\begin{align}
    &\int_{-\infty}^{+\infty} \phi(x) \diff x = 1,\quad \phi_{i,j}(x)= 2^{\frac{i}{2}} \phi(2^i x-j), \\ &\textrm{s.t.} \quad \langle \phi_{i,j}, \psi_{i',j'} \rangle = 0,\; i'>i,\;\forall j,j'  \nonumber
\end{align}
$\phi_{i,j}$ is designed to cover the scale up to $i$, hence the orthogonality requirement. The decomposition of $f(x)$ therefore becomes:
\begin{align}
    f(x) = \sum_{j=-\infty}^{+\infty} \langle \phi_{0,j}, f \rangle \phi_{0,j}(x) + \sum_{i=0}^{+\infty}\sum_{j=-\infty}^{+\infty} \langle \psi_{i,j}, f \rangle \psi_{i,j}(x) \label{eq:mra}
\end{align}
Note that although $i$ still goes to $+\infty$ in (\ref{eq:mra}), $i$ usually has an upper limit in practice since it is impossible to work with infinite frequency. 

In the $d$-dimensional case, we do not have a general orthogonal discrete $\R^d$ wavelet, unlike the continuous case. 
However, we can still perform discrete wavelet transform over each spatial dimension of the input, and we'd still be able to perfectly project and reconstruct the original function.
To be more specific, for sequential inputs $X \in \mathbb{R}^{n,d}$ of length $n$ and hidden dimension $d$, we will apply 2d wavelet transform over both the length and hidden dimension to generate the wavelet coefficients.

\subsection{Universal Approximation Power} \label{sec:random_feature_kernel}
\paragraph{Background about Attention}
Let $X \in \mathbb{R}^{n\times d}$ denotes the input sequence of length $n$ and hidden dimension $d$. 
A dense self-attention is shown below:
\begin{align}
    \mathrm{Attention}(X) &= \mathrm{Softmax}(\frac{Q K^\top}{\sqrt{d}}) V \label{eq:attn}
\end{align}
where $Q=X W_q$, $K=X W_k$, $V=X W_v$ with $W_q, W_k, W_v \in \mathbb{R}^{d \times m}$ stand for the query, key, and value, respectively.
The attention head size is denoted by $m$. 

In this subsection, we show that \paradigm can maintain the universal approximation power on seq-to-seq functions for Transformer and its variants. 
We illustrate this idea with proof for a slightly modified Performer~\cite{performer} under \paradigm.
The goal is to show that for any $f$ in $\mathcal{F}$, $\forall p \in [1, +\infty), \forall \epsilon >0$, we can find a $\bar{f}$ in the class of Performer-\paradigm, such that:
\begin{align*}
    d_p(f, \bar{f}) = \left( \int_{\R^{n\times d}} \|f(\mathbf{X}) - \bar{f}(\mathbf{X})\|^p_p \diff \mathbf{X} \right)^{\frac{1}{p}} \leq \epsilon
\end{align*}

We define the Performer-\paradigm class that has positional encoding, $h$ heads, head size $s$, hidden dimension $r$ as $\mathcal{W}^{h,s,r}$ with FAVOR+ kernel with an additional normalization on the input. 
\begin{theorem} \label{thm:main_theorem}
$\forall p \in [1, +\infty),\; \epsilon > 0$, and for any $f \in \mathcal{F}$, we can find a Performer-\paradigm network $\mathsf{w} \in \mathcal{W}^{2,1,4}$, such that $d_p(f,\mathsf{w}) \leq \epsilon$. 
\end{theorem}

The sketch of the proof is simple: since we have required the transformation pair to be invertible and exact, so for any seq-to-seq function, we can universally approximate it in the wavelet space and it is equivalent to having universal approximation power in the original space.
The detailed proof of Theorem~\ref{thm:main_theorem} is shown below.

\subsection{Proof for Theorem~\ref{thm:main_theorem}} \label{sec:appendix_proof}
We define the function class $\mathcal{F}$ to be the set of all countinous functions that map a compact domain in $\R^{n\times d}$ to $\R^{n\times d}$. 

We start from making the connection between random feature kernel and regular transformer block:
\begin{lemma} \label{lemma:favor}
(Asymptotic Result for FAVOR+) 
The following is true for independent random $w_i$,
\begin{align*}
    & \mathrm{MSE}(\hat{\mathrm{SM}}(x ,y))\\
    &=\frac{1}{m} \exp{(\|x+y\|^2)} \mathrm{SM}^2(x,y) (1-\exp{(-\|x+y\|^2)}) \\
    \Rightarrow &\lim_{\mathrm{SM}(x,y)\xrightarrow{} 0} \mathrm{MSE}(\hat{\mathrm{SM}}(x ,y)) \xrightarrow{} 0
\end{align*}
where $\mathrm{SM}$ denotes the softmax kernel, $\hat{\mathrm{SM}}$ denotes the random feature kernel, and $\mathrm{MSE}$ stands for mean-squared error.
\end{lemma}
The proof of this lemma can be found at~\cite[Lemma 2]{performer}. 
It tells us the the MSE error is upper bounded to a constant since $x$, $y$ is normalized beforehand, and vanishes to 0 as the original softmax kernel value tends to 0 and the number of random features $m$ tends to $+\infty$. 

Next we use the main theorem of~\cite{yun2019transformers}. 
We denote the transformer network class that has positional encoding, $h$ heads, head size $s$, and hidden dimension $r$ as $\mathcal{T}^{h,s,r}$. 
\begin{lemma} \label{lemma:transformer}
$\forall p \in [1, +\infty),\; \epsilon > 0$, and for any $f \in \mathcal{F}$, we can find a Transformer network $g\in \mathcal{T}^{2,1,4}$, such that $d_p(f,g) \leq \epsilon$. 
\end{lemma}
The proof of Lemma~\ref{lemma:transformer} constitutes of several steps, of which the first step is to approximate any function $f \in \mathcal{F}$ as a piece-wise constant function $\Tilde{f}$.  
Since $f$ is continuous, the piece-wise constant approximation can be of arbitrary accuracy.
Next they find a modified transformer $\Tilde{g}$ with hardmax operator and a special class of activations. Finally they show that the transformer block $g$ is able to approximate $g$. The functional distance is then bounded by:
\begin{align*}
    d_p(f,g) \leq d_p(f,\Tilde{f}) + d_p(\Tilde{f}, \Tilde{g}) + d_p(\Tilde{g}, g) \leq \epsilon 
\end{align*}
We show that with slight modification, the proof will work for Performer-\paradigm, and can be generalized to the \paradigm paradigm under certain constraints.

The proof is outlined below: 
For $\forall f \in \mathcal{F}$, its wavelet transform $\hat{f}$ (we will also use $f_w$ to denote this, see (\ref{eq:dwtf}) for details) is still continuous. 
Hence, the discretization claim remains valid. 
We can then effectively approximate the self-attention transformer block with the FAVOR+ block up to $\frac{\epsilon}{4}$ difference by controlling the number of random features $m$. In the end, the backward reconstruction is exact, the distance bound becomes when we control the other three terms to be less than $\frac{1}{4}\epsilon$ as well:
\begin{align*}
    & \phantom{{}\leq{}} d_p(f,\mathsf{w}) \\
    &\leq d_p(f_w,\Tilde{f}_w) + d_p(\Tilde{f}_w, \Tilde{g}) + d_p(\Tilde{g}, g) + d_p(g, \mathsf{w}) \\
    &\leq \epsilon \hspace{200pt minus 1fil} \qed \hfilneg
\end{align*}


\begin{table*}[t]
    \caption{Performance comparison of Transformers on Long Range Arena: Transformed spaces vs. Original space. We use $\mathcal{F}$/$\mathcal{W}$ to denote the Fourier space learning similar to AFNO~\cite{guibas2021adaptive} and GFNet~\cite{GFNET} or wavelet space (which adds an $O(n \log n)$/$O(n)$ complexity cost). We color the number green if it surpasses the baseline, red vice versa. Wavelet space ($\mathcal{W}$) demonstrated superior performance in 21 out of 25 architecture/task combinations compared to Fourier space.
    $\ssymbol{2}$ We reran Linformer \& Linear Attention for all (N/A, $\mathcal{F}, \mathcal{W}$) with the same additional five sets of hyperparameters because of convergence issues.$\ssymbol{3}$ We note that we are unable to reproduce a score close to the original Linformer performance on Pathfinder. $\ssymbol{4}$ This is the normalized version of Performer as described in Section~\ref{sec:random_feature_kernel}.}
    \vspace{-3mm}
  \begin{center}
    \small
    \renewcommand\tabcolsep{3pt}
    \resizebox{\textwidth}{!}{
    \begin{tabular}{cc | ccc | ccc | ccc | ccc | ccc}
    \toprule
    \multicolumn{2}{c}{\multirow{2}{*}{\textbf{Transformer Variants}}} & \multicolumn{3}{c}{\textbf{ListOps}} & \multicolumn{3}{c}{\textbf{Text}} & \multicolumn{3}{c}{\textbf{Retrieval}} & \multicolumn{3}{c}{\textbf{Image}} & \multicolumn{3}{c}{\textbf{Pathfinder}} \\
    \cmidrule(lr){3-5} \cmidrule(lr){6-8} \cmidrule(lr){9-11} \cmidrule(lr){12-14} \cmidrule(lr){15-17}
     & & \textbf{N/A} & $\mathcal{F}$ & $\mathcal{W}$ & \textbf{N/A} & $\mathcal{F}$ & $\mathcal{W}$ & \textbf{N/A} & $\mathcal{F}$ & $\mathcal{W}$ & \textbf{N/A} & $\mathcal{F}$ & $\mathcal{W}$ & \textbf{N/A} & $\mathcal{F}$ & $\mathcal{W}$ \\
    \midrule
    \textbf{Full} & $O(n^2)$ & 36.37 & \textcolor{purple}{17.80} & \textbf{\textcolor{ForestGreen}{37.15}} & 64.27 & \textcolor{purple}{56.42} & \textbf{\textcolor{ForestGreen}{74.82}} & 57.46 & \textcolor{purple}{51.78} & \textbf{\textcolor{ForestGreen}{72.43}} & 42.44 & \textcolor{purple}{31.41} & \textcolor{purple}{42.29} & 71.40 & \textcolor{purple}{50.55} & \textbf{\textcolor{ForestGreen}{78.25}} \\
    \textbf{Linformer} & $O(n)$ & 35.70 & \textcolor{ForestGreen}{36.15} & \textbf{\textcolor{ForestGreen}{37.65}} & 53.94 & \textbf{\textcolor{ForestGreen}{57.06}} & \textcolor{ForestGreen}{55.22} & 52.27 & \textcolor{ForestGreen}{55.93} & \textbf{\textcolor{ForestGreen}{65.85}} & 38.47$\ssymbol{2}$ & \textcolor{purple}{34.89}$\ssymbol{2}$ & \textbf{\textcolor{ForestGreen}{39.17}}$\ssymbol{2}$ & 66.44$\ssymbol{2}$ $\ssymbol{3}$ & \textcolor{purple}{61.76}$\ssymbol{2}$ & \textbf{\textcolor{ForestGreen}{70.21}}$\ssymbol{2}$ \\
    \textbf{Linear Att.} & $O(n)$ & 16.13 & \textbf{\textcolor{ForestGreen}{37.65}} & \textcolor{ForestGreen}{37.55} & 65.90 & \textcolor{ForestGreen}{71.66} & \textbf{\textcolor{ForestGreen}{71.93}} & 53.09 & \textbf{\textcolor{ForestGreen}{72.71}} & \textcolor{ForestGreen}{70.71} & 42.32$\ssymbol{2}$ & \textbf{\textcolor{ForestGreen}{51.07}}$\ssymbol{2}$ & \textcolor{purple}{40.83}$\ssymbol{2}$ & 75.91$\ssymbol{2}$ & \textcolor{purple}{70.45}$\ssymbol{2}$ & \textbf{\textcolor{ForestGreen}{76.43}}$\ssymbol{2}$ \\
    \textbf{Longformer} & $O(n)$ & 35.63 & \textcolor{purple}{18.95} & \textbf{\textcolor{ForestGreen}{36.65}} & 62.85 & \textcolor{purple}{55.36} & \textbf{\textcolor{ForestGreen}{74.99}} & 56.89 & \textcolor{purple}{52.52} & \textbf{\textcolor{ForestGreen}{66.21}} & 42.22 & \textcolor{purple}{29.12} & \textcolor{purple}{37.10} & 69.71 & \textcolor{purple}{50.38} & \textbf{\textcolor{ForestGreen}{78.15}} \\
    \textbf{Performer}$\ssymbol{4}$ & $O(n)$ & 18.01 & \textcolor{ForestGreen}{37.15} & \textbf{\textcolor{ForestGreen}{38.20}} & 65.40 & \textcolor{ForestGreen}{65.52} & \textbf{\textcolor{ForestGreen}{75.60}} & 53.82 & \textcolor{ForestGreen}{60.56} & \textbf{\textcolor{ForestGreen}{78.56}} & 42.77 & \textcolor{purple}{9.99} & \textbf{\textcolor{ForestGreen}{42.98}} & 77.05 & \textcolor{purple}{50.49} & \textbf{\textcolor{ForestGreen}{79.17}} \\
    \bottomrule
    \end{tabular}
    }
    \label{tab:main_appendix}
  \end{center}
  \vspace{-6mm}
\end{table*}
\begin{table}
  \begin{center}
    \caption{Evaluation results for the three adaptive parameterization schemes, we denote direct/orthogonal parameterization, and wavelet lifting as Ada/Ortho/Lift-\paradigm.}
    \vspace{1mm}
    \small
    \resizebox{\linewidth}{!}{
    \begin{tabular}{c|ccccc|cc} 
      \toprule
      Models & ListOps & Text & Retrieval & Image & Pathfinder & Avg & Avg (w/o r)\\
      \midrule
      \textbf{Transformer} & 36.37 & 64.27 & 57.46 & 42.44 & 71.40 & 54.39 & 53.62\\ 
      Ada\paradigm & \textbf{\textcolor{ForestGreen}{55.40}}	& \textcolor{ForestGreen}{81.60} & \textbf{\textcolor{ForestGreen}{79.27}} & \textbf{\textcolor{ForestGreen}{55.58}} &  \textcolor{ForestGreen}{81.12}  & \textbf{\textcolor{ForestGreen}{70.59}}	& \textbf{\textcolor{ForestGreen}{68.43}} \\
      Ortho\paradigm & \textcolor{ForestGreen}{45.95} & \textbf{\textcolor{ForestGreen}{81.63}} & \textcolor{ForestGreen}{71.52} & \textcolor{ForestGreen}{49.29} & \textcolor{ForestGreen}{81.13} & \textcolor{ForestGreen}{65.90} & \textcolor{ForestGreen}{64.50}\\
      Lift\paradigm & \textcolor{ForestGreen}{42.95} &	\textcolor{ForestGreen}{75.63} & \textcolor{purple}{56.45} & \textcolor{ForestGreen}{42.48} & \textbf{\textcolor{ForestGreen}{81.73}} & \textcolor{ForestGreen}{59.85}	& \textcolor{ForestGreen}{60.70} \\
      \midrule
      \textbf{Longformer} & 35.63 & 62.85 & 56.89 & 42.22 & 69.71 & 53.46 & 52.60 \\
      Ada\paradigm & \textbf{\textcolor{ForestGreen}{49.30}}	& \textbf{\textcolor{ForestGreen}{79.73}} & \textcolor{ForestGreen}{58.57} & \textbf{\textcolor{ForestGreen}{50.84}} &  \textbf{\textcolor{ForestGreen}{79.48}} & \textbf{\textcolor{ForestGreen}{63.66}} & \textbf{\textcolor{ForestGreen}{64.93}}\\
      Ortho\paradigm & \textcolor{ForestGreen}{39.45} & \textcolor{ForestGreen}{78.41} & \textbf{\textcolor{ForestGreen}{79.93}} & \textcolor{ForestGreen}{49.93} & \textcolor{ForestGreen}{79.47} & \textcolor{ForestGreen}{54.96} & \textcolor{ForestGreen}{54.96} \\
      Lift\paradigm & \textcolor{ForestGreen}{39.40} &	\textcolor{ForestGreen}{78.00} & \textcolor{purple}{53.27} & \textcolor{purple}{40.95} & \textcolor{ForestGreen}{75.80} & \textcolor{ForestGreen}{57.48} & \textcolor{ForestGreen}{58.54}\\
      \midrule
      \textbf{Linformer} & 35.70 & 53.94 & 52.27 & 38.47 & 66.44 & 49.36 & 48.64 \\
      Ada\paradigm & \textcolor{ForestGreen}{37.15}	& \textcolor{ForestGreen}{54.75} & \textcolor{ForestGreen}{61.09} & \textcolor{purple}{34.93} &  \textcolor{purple}{65.66} & \textcolor{ForestGreen}{50.72}	& \textcolor{purple}{48.12}\\
      Ortho\paradigm & \textbf{\textcolor{ForestGreen}{38.05}} & \textbf{\textcolor{ForestGreen}{56.93}} & \textcolor{ForestGreen}{60.25} & \textbf{\textcolor{ForestGreen}{39.45}} & \textcolor{purple}{65.35} & \textcolor{ForestGreen}{52.01} & \textbf{\textcolor{ForestGreen}{49.95}}\\
      Lift\paradigm & \textcolor{ForestGreen}{37.30} &	\textcolor{ForestGreen}{54.43} & \textbf{\textcolor{ForestGreen}{70.73}} & \textcolor{purple}{34.66} & \textcolor{purple}{63.49} & \textbf{\textcolor{ForestGreen}{52.12}} & \textcolor{purple}{47.47}\\
      \midrule
      \textbf{Linear Att.} & 16.13 & 65.90 & 53.09 & 42.32 & 75.91 & 50.67 & 50.06\\
      Ada\paradigm & \textcolor{ForestGreen}{38.90}	& \textcolor{ForestGreen}{76.82} & \textbf{\textcolor{ForestGreen}{71.38}} & \textbf{\textcolor{ForestGreen}{54.81}} &  \textbf{\textcolor{ForestGreen}{79.68}} & \textbf{\textcolor{ForestGreen}{64.32}}	& \textbf{\textcolor{ForestGreen}{62.55}}\\
      Ortho\paradigm & \textbf{\textcolor{ForestGreen}{39.55}} & \textbf{\textcolor{ForestGreen}{79.45}} & \textcolor{ForestGreen}{69.65} & \textcolor{ForestGreen}{49.93} & \textcolor{ForestGreen}{78.09} & \textcolor{ForestGreen}{55.86} & \textcolor{ForestGreen}{55.86}\\
      Lift\paradigm & \textcolor{ForestGreen}{38.35} &	\textcolor{ForestGreen}{73.39} & \textcolor{ForestGreen}{54.06} & \textcolor{ForestGreen}{44.39} & \textcolor{purple}{74.46} & \textcolor{ForestGreen}{56.93} & \textcolor{ForestGreen}{57.65} \\
      \midrule
      \textbf{Performer} & 18.01 & 65.40 & 53.82 & 42.77 & 77.05 & 51.41 & 50.81 \\
      Ada\paradigm & \textbf{\textcolor{ForestGreen}{46.05}}	& \textbf{\textcolor{ForestGreen}{80.93}} & \textbf{\textcolor{ForestGreen}{71.16}} & \textbf{\textcolor{ForestGreen}{52.06}} &  \textcolor{ForestGreen}{77.17} & \textbf{\textcolor{ForestGreen}{65.47}} & \textbf{\textcolor{ForestGreen}{64.05}}\\
      Ortho\paradigm & \textcolor{ForestGreen}{39.80} & \textcolor{ForestGreen}{79.10} & \textcolor{ForestGreen}{57.67} & \textcolor{ForestGreen}{48.78} & \textbf{\textcolor{ForestGreen}{78.09}} & \textcolor{ForestGreen}{60.69}	& \textcolor{ForestGreen}{61.44} \\
      Lift\paradigm & \textcolor{ForestGreen}{39.85} & \textcolor{ForestGreen}{75.96} & \textcolor{purple}{52.75} & \textcolor{purple}{39.97} & \textcolor{purple}{76.20} & \textcolor{ForestGreen}{56.95} & \textcolor{ForestGreen}{58.00} \\
      \bottomrule
    \end{tabular}
    }
    
    \label{tab:adaptive_table_full}
  \end{center}
  \vspace{-6mm}
\end{table}

\subsection{LRA Configuration Details} \label{sec:appendix_lra}
We tried to follow all hyperparameters as suggested for each of the attention approximations with exceptions on Linformer and Linear Trans. in Image and Pathfinder. For them, we experimented with five additional configurations as shown in Table~\ref{tab:lra_hyper}. 

For all fixed wavelet transform conducted in this work, we use Daubechies-2~\cite{daubechies1992ten} as the basis and we set the level of decomposition to 1.

For Performer, the number of random features in the random feature kernel is set as 256 for all text tasks (ListOps, Text, Retrivial), 512 for all image tasks (Image, Pathfinder).

\begin{table}[t]
    \centering
    \small
    \caption{Additional hyperparameter configurations tried for Linformer and Linear Att. in Image and Pathfinder}
    \begin{tabular}{l | ccccc }
    \toprule
    \textbf{Hyperparameter} & Config$_1$ & Config$_2$ & Config$_3$ & Config$_4$ & Config$_5$ \\
    \midrule
    \textbf{Layers} & 1 & 1 & 2 & 2 & 2  \\
    \textbf{Embedding Dim.} & 128 & 128 & 128 & 256 & 256 \\
    \textbf{Attention Dim.} & 64 & 64 & 64 & 64 & 64 \\
    \textbf{MLP Dim.} & 128 & 128 & 256 & 1024 & 512 \\
    \textbf{Attention Heads} & 8 & 8 & 2 & 4 & 4 \\
    \textbf{Dropout} & 0.2 & 0.1 & 0.1 & 0.1 & 0.2 \\
    \textbf{Attention Dropout} & 0.1 & 0.1 & 0.1 & 0.1 & 0.1  \\
    \bottomrule
    \end{tabular}
    
    \label{tab:lra_hyper}
\end{table}
\begin{table}[t]
    \centering
    \small
    \caption{Hyperparameter configurations for parameterized \paradigm experiments.}
    \begin{tabular}{l | ccccc }
    \toprule
    \textbf{Hyperparameter} &  ListOps & Text & Retrieval & Image & Pathfinder \\
    \midrule
    \textbf{Batch Size} & 400 & 128 & 64 & 64 & 512  \\
    \textbf{Max Step} & 80k & 50k & 50k & 200k & 500k  \\
    \textbf{Min Step} & 5k & 20k & 20k & 20k & 20k \\
    \textbf{Layers} & 8 & 6 & 6 & 8 & 1  \\
    \textbf{Embedding Dim.} & 128 & 256 & 256 & 128 & 128 \\
    \textbf{Attention Dim.} & 64 & 256 & 128 & 64 & 64 \\
    \textbf{MLP Dim.} & 128 & 1024 & 256 & 128 & 128 \\
    \textbf{Attention Heads} & 1 & 1 & 1 & 1 & 8 \\
    \textbf{Ada\paradigm WLen.} & 40 & 16 & 40 & 40 & 40 \\
    \textbf{Ortho\paradigm WLen.} & 16 & 16 & 16 & 16 & 16 \\
    \textbf{Lift\paradigm Lev.} & 3 & 3 & 3 & 3 & 3 \\
    \textbf{Dropout} & 0.1 & 0.1 & 0.1 & 0.1 & 0.2  \\
    \textbf{Attention Dropout} & 0.1 & 0.1 & 0.1 & 0.1 & 0.1  \\
    \textbf{Readout} & CLS &  CLS & CLS & CLS2 & MEAN  \\
    \bottomrule
    \end{tabular}
    
    \label{tab:parameterization_config}
\end{table}

We use the same set of hyperparameters for all the attention methods on individual tasks, the detailed setting is shown in Table~\ref{tab:parameterization_config}. We adjust the training length to stabilize the adaptive wavelets, the training will take over min steps and then will terminate with patience ($=10\%*\text{max step}$). We also find out that on the image task, it is better to use both the first and last output as the readout (CLS2) for parameterized wavelet transformation.

\begin{table}
  \begin{center}
    \caption{Evaluation results on Long-Range Arena benchmark. We show both the average accuracy (Avg) and average accuracy without Retrieval (Avg (w/o r)) since LUNA 256, Nystr\"omformer, and our \method coupled with full attention and direct wavelet parameterization (\methodname) all use prolonged training steps on Retrieval.} 
    \small
    \resizebox{\linewidth}{!}{
    \begin{tabular}{l|ccccc| cc} 
      \toprule
      Model & ListOps & Text & Retrieval & Image & Pathfinder & Avg & Avg (w/o r)\\
      \midrule
      Transformer & 36.37 & 64.27 & 57.46 & 42.44 & 71.40 & 54.39 & 53.62 \\
      \midrule
      Local Attention & 15.82 & 52.98 & 53.39 & 41.46 & 66.63 & 46.06 & 44.22 \\
      Sparse Trans. & 17.07 & 63.58 & 59.59 & 44.24 & 71.71 & 51.24 & 49.15 \\
      Longformer & 35.63 & 62.85 & 56.89 & 42.22 & 69.71 & 53.46 & 52.60 \\
      Linformer & 35.70 & 53.94 & 52.27 & 38.56 & 76.34 & 51.36 & 51.14 \\
      Reformer & 37.27 & 56.10 & 53.40 & 38.07 & 68.50 & 50.67 & 49.99 \\ 
      Sinkhorn Trans. & 33.67 & 61.20 & 53.83 & 41.23 & 67.45 & 51.39 & 50.89\\
      Synthesizer & 36.99 & 61.68 & 54.67 & 41.61 & 69.45 & 52.88 & 52.43\\
      BigBird & 36.05 & 64.02 & 59.29 & 40.83 & 74.87 & 55.01 & 53.94 \\
      Linear Trans. & 16.13 & 65.90 & 53.09 & 42.34 & 75.30 & 50.55 & 49.92\\
      Performer & 18.01 & 65.40 & 53.82 & 42.77 & 77.05 & 51.41 & 50.81\\
      FNet & 35.33 & 65.11 & 59.61 & 38.67 & 77.80 & 55.30 & 54.23 \\
      LUNA 256 & 37.98 & 65.78 & \textbf{79.56} & 47.86 & 78.55 & 61.95 & 57.54 \\
      Nystr\"omformer & 37.15 & 65.52 & \textbf{79.56} & 41.58 & 70.94 & 58.95 & 53.80 \\
      \midrule
      \methodname & \textbf{55.40} & \textbf{81.60} & 79.27 & \textbf{55.58} & \textbf{81.12} & \textbf{70.59} & \textbf{68.42}\\
      \bottomrule
    \end{tabular}
    }
    
    \label{tab:lra_table}
  \end{center}
\end{table}
\begin{table}
\begin{center}
\small
\caption{Training, inference latency, and the number of additional wavelet parameters in parametrized wavelet transform for fixed and adaptive \paradigm, where $d$ denotes the hidden dimension.}
\resizebox{\linewidth}{!}{
\begin{tabular}{c|ccccc}
\toprule
Schemes                    & Transformer & Fixed Daubechies-2 & Ada\paradigm & Ortho\paradigm & Lift\paradigm \\
\midrule
Avg Train Latency        & 100.00\%    & 102.56\%           & 112.43\%              & 119.97\%                & 63.52\%                \\
Avg Inference Latency    & 100.00\%    & 103.03\%           & 112.16\%              & 119.38\%                & 66.61\%                \\
Add. Wavelet Parameters & 0 & 0 &  Wavelet Length * $d$ & (Wavelet Length / 2) * $d$	 & 0 \\
\bottomrule
\end{tabular}
}
\label{tab:cost_table}
\end{center}
\end{table}

\subsection{Ablation Study} \label{sec:ablation}
We conduct an ablation study for \method, as shown in Table~\ref{tab:ablation}. For (Linear), We limit the transformation in wavelet space to be linear. For (Fourier), we use the Fourier transform as the transformation mechanism for \method. For (Forward Fourier), we only use the forward Fourier transform without backward transform.
It can be observed that performance dropped significantly in all cases, indicating the necessity of non-linearity in wavelet space and forward-backward wavelet transform. 
\begin{table}
    \centering
    \caption{Ablation study on Long-Range Arena benchmark.}    \label{tab:ablation}
    \resizebox{\linewidth}{!}{
    \begin{tabular}{c|ccccc| cc} 
      \toprule
      Model & ListOps & Text & Retrieval & Image & Pathfinder & Avg & Avg (w/r)\\
      \midrule
      \method & 38.20 & 75.60 & 78.56 & 42.98 & 79.17 & 62.90 & 58.99\\
      \midrule
      Linear & 37.70 & 55.36 & 55.27 & 15.75 & 50.58 & 42.93 & 39.84 \\
      Fourier & 36.85 & 65.52 & 60.56 & 9.99 & 50.49 & 44.68 & 40.71 \\ 
      Forward Fourier & 37.15 & 64.91 & 65.98 & 37.84 & 53.39 & 51.85 & 48.32 \\
      \bottomrule
    \end{tabular}
   }
\end{table}

For fixed \paradigm, we also try out different wavelet families and decomposition level when paired with Performer, results shown in Table~\ref{tab:ablation_level_family}. 
\begin{table}
    \centering
    \caption{Ablation study for different wavelet families \& decomposition levels for fixed Performer-\paradigm on Long-Range Arena benchmark.}    \label{tab:ablation_level_family}
    \begin{tabular}{c|ccccc} 
      \toprule
      Config & ListOps & Text & Retrieval & Image & Pathfinder \\
      \midrule
      Daubechies-2, L=1 & 38.20 & 75.60 & 78.56 & 42.98 & 79.17 \\
      \midrule
      Daubeuchies-3, L=1 & 37.85 & 76.86 & 73.62 & 42.30 & 78.30\\
      Daubeuchies-3, L=2 & 37.40 & 76.84 & 75.43 & 41.20 & 50.52\\ 
      Coiflet-1, L=1 & 36.85 & 76.72 & 75.43 & 41.42 & 50.58\\
      Coiflet-1, L=2 & 37.65 & 75.97 & 75.29 & 43.2 & 77.49\\
      Symlet-2, L=1 & 37.50 & 75.07 & 75.59 & 42.85 & 49.85\\
      Symlet-2, L=2 & 37.45 & 75.63 & 74.51 & 41.03 & 77.39\\
      Symlet-2, L=3 & 37.55 & 75.39 & 76.24 & 40.88 & 76.84\\
      \bottomrule
    \end{tabular}
\end{table}

We further test the necessity of wavelets in direct wavelet parameterization scheme. We tried two other initializations on ListOps task when coupled with full attention, one with random Gaussian initialization $\mathrm{N}(\mu=0, \sigma=0.02)$, the other one with damped sinusoidal wave initialization. It can be observed from Table~\ref{tab:ablation_init} that both alternative initializations induced significant performance deterioration.
\begin{table}
\centering
    \caption{Ablation study for Daubechies wavelet initialization, Gaussian initialization, and damped sinusoidal wave initialization. All experiments are using full attention as the non-linearity.}    \label{tab:ablation_init}
    \begin{tabular}{c|c} 
      \toprule
       Initialization & ListOps \\
      \midrule
      Daubechies init. & 55.4 \\ 
      \midrule
      Gaussian init. &  45.9 \\
      Damped Cos init. &  44.65 \\ 
      \bottomrule
    \end{tabular}
\end{table}

\subsection{Model Runtime Analysis} \label{sec:appendix_runtime}
Since our framework involves two additional steps in attention computation, the forward and backward wavelet transform, it is important to measure the add-on cost of these two transformations. We show the training \& inference latency, and the number of additional parameters used in adaptive wavelets in Table~\ref{tab:cost_table}. The run-time data is collected for 2,000 steps on the Text dataset with sequence length being 4,096. All \paradigm variants pay a small overhead (linear to sequence length) on the wavelet transformations but also gain an advantage in efficiency due to the halved lengths in each decomposition level. For example, with wavelet lifting, we observe a 40\% less latency due to the higher decomposition level (L=3) and simpler decomposition scheme. To illustrate the decomposition effect, an input of length 4096 with three decomposition levels will be transformed into four sequences with lengths 2048, 1024, 512, 512, thus delivering a speed-up ($2048^2 + 1024^2 + 512^2 + 512^2 \ll 4096^2$).

\subsection{Exploratory Study on Learned Wavelets} \label{sec:exploratory}
Since we have trained the adaptive wavelet in the end-to-end fashion, we are naturally drawn to this question: what kind of wavelets has been learned? We use the best-performing direct parameterization scheme as an example, to empirically examine the learned wavelets.

We initiate our analysis from one commonly studied property of wavelets that characterizes the phase response of the wavelet~\cite{rieder1998parameterization} -- symmetry. A closer-to-symmetric wavelet (such as symlet) will have a closer-to-linear phase response, and a less symmetric wavelet (such as Daubechies) will have a more distorted phase response. 
We measure the symmetry by calculating the $\ell_1$ norm between the unit-normalized wavelet and the wavelet's transpose. A perfectly symmetric wavelet will have 0 on this measure.

\begin{figure}
\centering
\includegraphics[width=0.78\columnwidth]{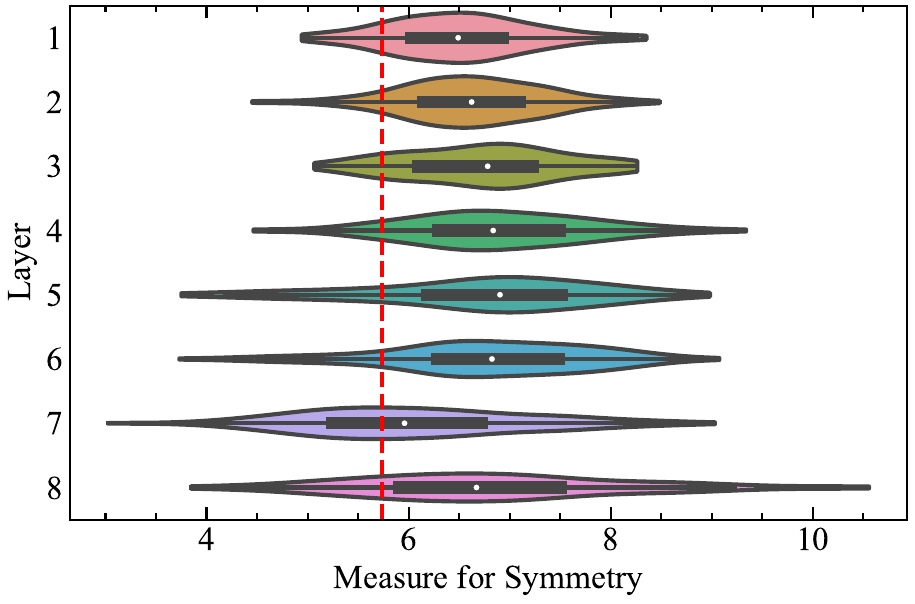}
\caption{We show the measure for symmetry for all the layers of the \methodname trained on ListOps task. The density plot of each layer shows the distribution for all wavelets of each individual hidden dimension (in this case 128 wavelets). The red dotted vertical line denotes the measure for the Daubechies-20 wavelet, which is the wavelets' initialization value.}
\label{fig:non_sym_adawise}
\end{figure}

From Figure~\ref{fig:non_sym_adawise}, we observe that the variance in symmetry grows larger when going from shallow to deep layers. Also on average, the learned wavelets are almost always less symmetric compared to the Daubechies-20 wavelet. These results indicate that it is important to turn wavelet transformation into an adaptive process since the optimal wavelet design varies across the layers and the hidden dimensions.

\end{document}